\documentclass[10pt,twocolumn,letterpaper]{article}

\usepackage{cvpr}
\usepackage{times}
\usepackage{epsfig}
\usepackage{graphicx}
\usepackage{amsmath}
\usepackage{amssymb}
\usepackage{float}
\usepackage{bbm}
\usepackage{color}

\DeclareMathOperator*{\argmin}{\arg\!\min}
\DeclareMathOperator*{\argmax}{\arg\!\max}

\usepackage[pagebackref=true,breaklinks=true,letterpaper=true,colorlinks,bookmarks=false]{hyperref}

\cvprfinalcopy % *** Uncomment this line for the final submission

\def\cvprPaperID{3350} % *** Enter the CVPR Paper ID here
\def\httilde{\mbox{\tt\raisebox{-.5ex}{\symbol{126}}}}

\ifcvprfinal\fi
\begin{document}

   %\newcommand{\comment}[1]{}

% new definition, clean it when I have time
\def\httilde{\mbox{\tt\raisebox{-.5ex}{\symbol{126}}}}
\def\balpha{\mbox{\boldmath $\alpha$}}
\def\bDelta{{\bf \Delta}}
\def\bLambda{{\bf \Lambda}}
\def\bvarphi{{\bf \varphi}}
\def\bTheta{{\bf \Theta}}
\def\btheta{\mbox{\boldmath $\theta$}}

\def\bPhi{\mbox{\boldmath{$\Phi$}}}
\def\vbphi{\vec{\mbox{\boldmath $\phi$}}}
\def\bb{{\bf b}}
\def\h{{\bf h}}
\def\bd{{\bf d}}
\def\ba{{\bf a}}
\def\bc{{\bf c}}
\def\p{{\bf p}}
\def\e{{\bf e}}
\def\s{{\bf s}}
\def\X{{\bf X}}
\def\x{{\bf x}}
\def\Y{{\bf Y}}
\def\y{{\bf y}}
\def\K{{\bf K}}
\def\k{{\bf k}}
\def\p{{\bf p}}
\def\bc{{\bf c}}
\def\A{{\bf A}}
\def\B{{\bf B}}
\def\C{{\bf C}}
\def\V{{\bf V}}
\def\S{{\bf S}}
\def\T{{\bf T}}
\def\W{{\bf W}}
\def\I{{\bf I}}
\def\U{{\bf U}}
\def\g{{\bf g}}
\def\G{{\bf G}}
\def\Q{{\bf Q}}
\def\d{{\bf d}}
\def\eg{{\it e.g.}}
\def\etal{{\it et. al}}
\def\H{{\bf H}}
\def\cR{{\bf R}}
\def\J{{\bf J}}
\def\bt{{\bf t}}
\def\bv{{\bf v}}
\def\R{{\bf R}}

\def\balpha{\mbox{\boldmath $\alpha$}}
\def\bdelta{\mbox{\boldmath $\delta$}}
\def\bzeta{\mbox{\boldmath $\zeta$}}
\def\bphi{\mbox{\boldmath $\phi$}}
\def\btau{\mbox{\boldmath $\tau$}}
\def\bmu{\mbox{\boldmath $\mu$}}
\def\bsigma{\mbox{\boldmath $\sigma$}}
\def\bSigma{{\bm \Sigma} }
\def\btheta{\mbox{\boldmath $\theta$}}
\def\dbphi{\dot{\mbox{\boldmath $\phi$}}}
\def\dbtau{\dot{\mbox{\boldmath $\tau$}}}
\def\dbtheta{\dot{\mbox{\boldmath $\theta$}}}
\def\bGamma{\mbox{\boldmath $\Gamma$}}
\def\bDelta{\mbox{\boldmath $\Delta$}}
\def\blambda{\mbox{\boldmath $\lambda $}}
\def\bOmega{\mbox{\boldmath $\Omega $}}
\def\bbeta{\mbox{\boldmath $\beta $}}
\def\bupsilon{\mbox{\boldmath $\Upsilon$}}
\def\myphi{\phi}
\def\bPhi{\mbox{\boldmath{$\Phi$}}}
\def\bLambda{\mbox{\boldmath{$\Lambda$}}}
\def\bSigma{\mbox{\boldmath{$\Sigma$}}}

\def\balpha{\mbox{\boldmath{$\alpha$}}}
\def\bbeta{\mbox{\boldmath{$\beta$}}}
\def\bdelta{\mbox{\boldmath{$\delta$}}}
\def\bgamma{\mbox{\boldmath{$\gamma$}}}
\def\blambda{\mbox{\boldmath{$\lambda$}}}
\def\bsigma{\mbox{\boldmath{$\sigma$}}}
\def\btheta{\mbox{\boldmath{$\theta$}}}
\def\bomega{\mbox{\boldmath{$\omega$}}}
\def\bxi{\mbox{\boldmath{$\xi$}}}
%%%%

\def\bigO2{\mbox{${\cal O}$}}
\def\bigO{O}

\newcommand{\bH}{\mathbf{H}}
\def\mA{\mathcal{A}}
\def\mB{\mathcal{B}}
\def\mC{\mathcal{C}}
\def\mD{\mathcal{D}}
\def\mG{\mathcal{G}}
\def\mV{\mathcal{V}}
\def\mE{\mathcal{E}}
\def\mF{\mathcal{F}}
\def\mH{\mathcal{H}}
\def\mL{\mathcal{L}}
\def\mM{\mathcal{M}}
\def\mN{\mathcal{N}}
\def\mK{\mathcal{K}}
\def\mR{\mathcal{R}}
\def\mS{\mathcal{S}}
\def\mT{\mathcal{T}}
\def\mU{\mathcal{U}}
\def\mW{\mathcal{W}}
\def\mX{\mathcal{X}}
\def\mY{\mathcal{Y}}
\def\1n{\mathbf{1}_n}
\def\0{\mathbf{0}}
\def\1{\mathbf{1}}
\def\etal{{\em et al.}}

\def\balpha{\mbox{\boldmath $\alpha$}}
\def\bdelta{\mbox{\boldmath $\delta$}}
\def\bzeta{\mbox{\boldmath $\zeta$}}
\def\bphi{\mbox{\boldmath $\phi$}}
\def\btau{\mbox{\boldmath $\tau$}}
\def\bmu{\mbox{\boldmath $\mu$}}
\def\bsigma{\mbox{\boldmath $\sigma$}}
\def\bSigma{{\bm \Sigma} }
\def\btheta{\mbox{\boldmath $\theta$}}
\def\dbphi{\dot{\mbox{\boldmath $\phi$}}}
\def\dbtau{\dot{\mbox{\boldmath $\tau$}}}
\def\dbtheta{\dot{\mbox{\boldmath $\theta$}}}
\def\bGamma{\mbox{\boldmath $\Gamma$}}
\def\bDelta{\mbox{\boldmath $\Delta$}}
\def\blambda{\mbox{\boldmath $\lambda $}}
\def\bOmega{\mbox{\boldmath $\Omega $}}
\def\bbeta{\mbox{\boldmath $\beta $}}
\def\bupsilon{\mbox{\boldmath $\Upsilon$}}
\def\myphi{\phi}
\def\bPhi{\mbox{\boldmath{$\Phi$}}}
\def\bLambda{\mbox{\boldmath{$\Lambda$}}}
\def\bSigma{\mbox{\boldmath{$\Sigma$}}}

\def\balpha{\mbox{\boldmath{$\alpha$}}}
\def\bbeta{\mbox{\boldmath{$\beta$}}}
\def\bdelta{\mbox{\boldmath{$\delta$}}}
\def\bgamma{\mbox{\boldmath{$\gamma$}}}
\def\blambda{\mbox{\boldmath{$\lambda$}}}
\def\bsigma{\mbox{\boldmath{$\sigma$}}}
\def\btheta{\mbox{\boldmath{$\theta$}}}
\def\bomega{\mbox{\boldmath{$\omega$}}}
\def\bxi{\mbox{\boldmath{$\xi$}}}

\def\bPsi{\mbox{\boldmath $\Psi $}}
\def\bone{\mbox{\bf 1}}
\def\bzero{\mbox{\bf 0}}

\def\WB{{\bf WB}}

\def\A{{\bf A}}
\def\B{{\bf B}}
\def\C{{\bf C}}
\def\D{{\bf D}}
\def\E{{\bf E}}
\def\F{{\bf F}}
\def\G{{\bf G}}
\def\H{{\bf H}}
\def\I{{\bf I}}
\def\J{{\bf J}}
\def\K{{\bf K}}
\def\L{{\bf L}}
\def\M{{\bf M}}
\def\N{{\bf N}}
\def\O{{\bf O}}
\def\P{{\bf P}}
\def\Q{{\bf Q}}
\def\R{{\bf R}}
\def\S{{\bf S}}
\def\T{{\bf T}}
\def\U{{\bf U}}
\def\V{{\bf V}}
\def\W{{\bf W}}
\def\X{{\bf X}}
\def\Y{{\bf Y}}
\def\Z{{\bf Z}}

\def\b{{\bf b}}
\def\bc{{\bf c}}
\def\bd{{\bf d}}
\def\e{{\bf e}}
\def\f{{\bf f}}
\def\g{{\bf g}}
\def\h{{\bf h}}
\def\i{{\bf i}}
\def\j{{\bf j}}
\def\k{{\bf k}}
\def\l{{\bf l}}
\def\m{{\bf m}}
\def\n{{\bf n}}
\def\o{{\bf o}}
\def\p{{\bf p}}
\def\q{{\bf q}}
\def\br{{\bf r}}
\def\s{{\bf s}}
\def\t{{\bf t}}
\def\u{{\bf u}}
\def\v{{\bf v}}
\def\w{{\bf w}}
\def\bx{{\bf x}}
\def\y{{\bf y}}
\def\z{{\bf z}}

\def\vbphi{\vec{\mbox{\boldmath $\phi$}}}
\def\vbtau{\vec{\mbox{\boldmath $\tau$}}}
\def\vbtheta{\vec{\mbox{\boldmath $\theta$}}}
\def\vI{\vec{\bf I}}
\def\vR{\vec{\bf R}}
\def\vV{\vec{\bf V}}

%%% Vector notation for sections 3 and 4
%%% Vector notation for sections 3 and 4
\def\mvec{\vec{m}}
\def\fvec{\vec{f}}
\def\appfvec{\vec{f}_k}
\def\avec{\vec{a}}
\def\bvec{\vec{b}}
\def\evec{\vec{e}}
\def\uvec{\vec{u}}
\def\xvec{\vec{x}}
\def\wvec{\vec{w}}
\def\gradvec{\vec{\nabla}}

\def\aM{\mbox{\bf a}_M}
\def\aS{\mbox{\bf a}_S}
\def\aO{\mbox{\bf a}_O}
\def\aL{\mbox{\bf a}_L}
\def\aP{\mbox{\bf a}_P}
\def\ai{\mbox{\bf a}_i}
\def\aj{\mbox{\bf a}_j}
\def\an{\mbox{\bf a}_n}
\def\a1{\mbox{\bf a}_1}
\def\a2{\mbox{\bf a}_2}
\def\a3{\mbox{\bf a}_3}
\def\a4{\mbox{\bf a}_4}

\def\sx{\mbox{\scriptsize\bf x}}
\def\st{\mbox{\scriptsize\bf t}}
\def\ss{\mbox{\scriptsize\bf s}}
\def\cR{{\cal R}}
\def\calD{{\cal D}}
\def\calS{{\cal S}}

\def\sigmae{\sigma}
\def\sigmam{\sigma}

\def\balpha{\mbox{\boldmath{$\alpha$}}}
\def\bbeta{\mbox{\boldmath{$\beta$}}}
\def\bdelta{\mbox{\boldmath{$\delta$}}}
\def\bgamma{\mbox{\boldmath{$\gamma$}}}
\def\blambda{\mbox{\boldmath{$\lambda$}}}
\def\bsigma{\mbox{\boldmath{$\sigma$}}}
\def\btheta{\mbox{\boldmath{$\theta$}}}
\def\bomega{\mbox{\boldmath{$\omega$}}}
\def\bxi{\mbox{\boldmath{$\xi$}}}

\def\dx{{\delta \x}}
\def\dref{{\d_{ref}}}
\def\px{{\partial \x}}
\def\fxp{\f(\x, \p)}

\def\dfp{\mathbf{d}(\mathbf{f}(\x,\mathbf{p}))}
\def\dfpk{\mathbf{d}(\mathbf{f}(\x,\mathbf{p}^k))}
\def\Ep{E(\mathbf{\d, \p})}
\newcommand{\deltap}[1]{\Delta^{#1}}
\newcommand{\Jp}[1]{\J^{#1}}
\newcommand{\Hp}[1]{\H^{#1}}
\newcommand{\Hpnewton}[1]{\H^{#1}_{nt}}
\newcommand{\Psip}[1]{\Psi^{#1}}
\newcommand{\Phip}[1]{\Phi^{#1}}
\newcommand{\dPsip}[1]{\Psi_{#1}}
\newcommand{\dPhip}[1]{\Phi_{#1}}

\newcommand{\dn}{\d_{s}}
\newcommand{\dc}{\d}
\newcommand{\dnxt}[1]{\dn(\f(\x, #1))}
\newcommand{\dcur}[1]{\dc(\f(\x, #1))}

\newcommand{\one}{\mathbf{1}}
\newcommand{\zero}{\mathbf{0}}
\newcommand{\real}{\mathbb{R}}

\newcommand{\denselist}{\itemsep -1pt}
\newcommand{\sparselist}{\itemsep 1pt}

%%%%%%%%% TITLE
\title{Deep Unsupervised Similarity Learning using Partially Ordered Sets}

\author{Miguel A. Bautista\thanks{Both authors contributed equally to this work.} \ , Artsiom Sanakoyeu\footnotemark[1] \ , Bj{\"o}rn Ommer\\
Heidelberg Collaboratory for Image Processing\\
IWR, Heidelberg University, Germany\\ \texttt{firstname.lastname@iwr.uni-heidelberg.de}}
\maketitle

%%%%%%%%% ABSTRACT
\begin{abstract}

Unsupervised learning of visual similarities is of paramount importance to computer vision, particularly due to lacking training data for fine-grained similarities. Deep learning of similarities is often based on relationships between pairs or triplets of samples. Many of these relations are unreliable and mutually contradicting, implying inconsistencies when trained without supervision information that relates different tuples or triplets to each other. To overcome this problem, we use local estimates of reliable (dis-)similarities to initially group samples into compact surrogate classes and use local partial orders of samples to classes to link classes to each other. Similarity learning is then formulated as a partial ordering task with soft correspondences of all samples to classes. Adopting a strategy of self-supervision, a CNN is trained to optimally represent samples in a mutually consistent manner while updating the classes.  The similarity learning and grouping procedure are integrated in a single model and optimized jointly. The proposed unsupervised approach shows competitive performance on detailed pose estimation and object classification. 

\end{abstract}

%%%%%%%%% BODY TEXT
\section{Introduction}

% \item Visual similarity learning is the foundation for numerous computer vision subtasks. Motivation for unsupervised learning on a different paragraph for emphatization.

Visual similarities lie at the heart of a large number of computer vision tasks ranging from low-level image processing to high-level understanding of human poses or object classification. Of the numerous techniques for similarity learning, supervised methods have been a popular technique, leading to formulations in which similarity learning was casted as a ranking \cite{simlearning1}, regression \cite{simregression}, and classification \cite{videoparsing} task. In recent years, with the advent of Convolutional Neural Networks (CNN), formulations based on a ranking (i.e. ordering) of pairs or triplets of samples according to their similarity have shown impressive results \cite{ConvNetpretext2}. However, to achieve this  performance boost, these CNN architectures require millions of samples of supervised training data or at least the fine-tuning \cite{ConvNetpretext1} on large datasets such as PASCAL VOC.

Although the amount of accessible image data is growing at an ever increasing rate, supervised labeling of similarities is very costly. In addition, not only similarities between images are important, but especially between objects and their parts.  Annotating the fine-grained similarities between all these entities is a futile undertaking, in particular for the large-scale datasets typically used for training CNNs. Deep unsupervised learning of similarities is, therefore, of great interest to the vision community, since it does not require any labels for pre-training or fine-tuning. In this way we can utilize large image datasets without being limited by the need for costly manual annotations.
%ToDo:similar to nips? revise

\begin{figure}[!t]
\centering
\includegraphics[width=0.48\textwidth]{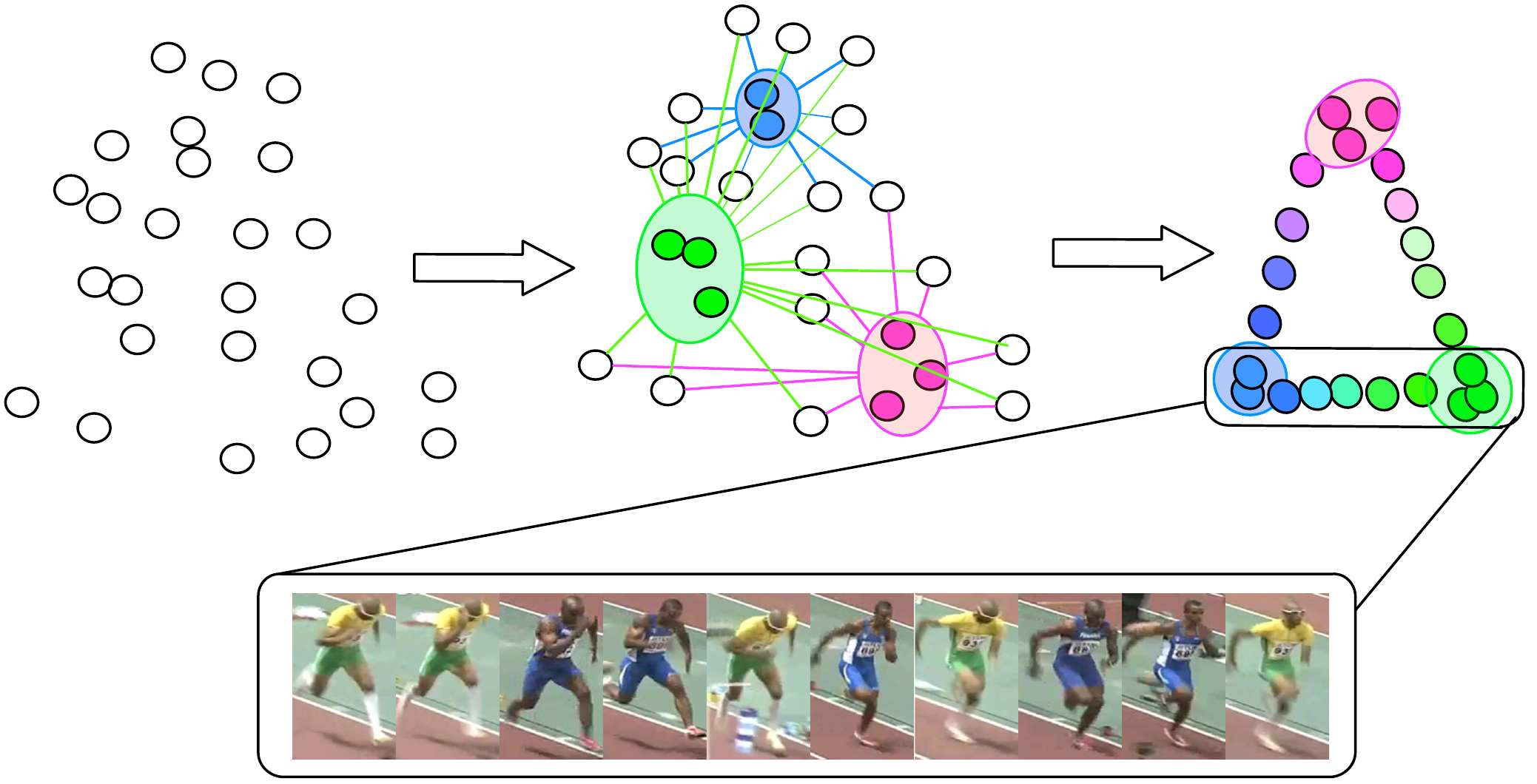}
\caption{Visualization of the interaction between surrogate classes and partially ordered sets (posets). Our approach starts with a set of unlabeled samples, building small surrogate classes and generating posets to unlabeled samples to learn fine-grained similarities.}
\label{fig:intro}
\end{figure}

To utilize the vast amounts of available unlabeled training data, there is a quest to leverage context information intrinsic to images/video for \textit{self-}supervision. However, this context is typically highly local (i.e position of patches in the same image \cite{ConvNetpretext1}, object tracks through short number of frames \cite{ConvNetpretext2} or image impainting \cite{impainting}), establishing relations between tuples \cite{ConvNetpretext1} or triplets \cite{shuffleandlearn,Yang_2016_CVPR,ConvNetpretext2} of images. Hence, these approaches utilize loss functions that order a positive $I_p$ and a negative $I_n$ image with respect to an anchor image $I_a$ so that, $d(I_a,I_p)<d(I_a,I_n)$. During training, these methods rely on the CNN to indirectly learn comparisons between samples that were processed in independent training batches, and generalize to unseen data.

Instead of relying on the CNN to indirectly balance and learn sample comparisons unseen during training, a more natural approach is to explicitly encode richer relationships between samples as supervision. In this sense, an effective approach to tackle unsupervised similarity learning is to frame it as a series of surrogate (i.e. artificially created) classification tasks \cite{exemplarcnn,cliquecnn}. Therefore, mutually similar samples are assigned the same class label, otherwise a different label. To obtain surrogate classification tasks, compact groups of mutually similar samples are computed by clustering \cite{cliquecnn} over a weak initial representation (e.g standard features such as HOG). Then, each group receives a mutually exclusive label and a CNN is trained to solve the associated classification problem, thereby learning a representation that encodes similarity in the intermediate layers. However, given the unreliability of initial similarities, a large number of training samples are neither mutually similar nor dissimilar and are, thus, not assigned to any of the compact surrogate classes. Consequentially they are ignored during training, hence overlooking important information. Also, classification can yield fairly coarse similarities, considering the discrete nature of the classes. Furthermore, the similarities learnt by the different classification tasks are not optimized jointly, which can lead to mutually contradicting relationships, since transitivity is not captured.

\begin{figure*}[ht]
\includegraphics[width=\textwidth, height= 2.7 cm]{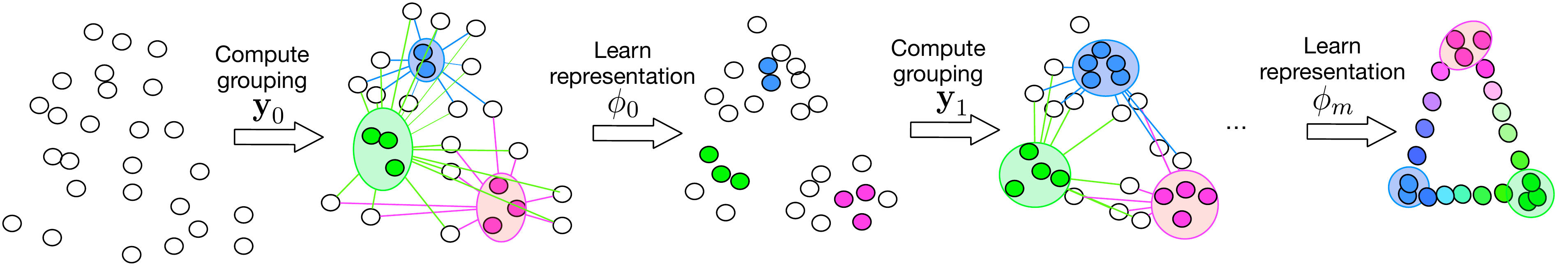}
\caption{Visual summary of our approach. In the $\y$-steps the clustering procedure computes surrogate classes (shaded in color) based on the current representation. In the $\phi$-steps we learn a representation using the surrogate classes and partial orders of samples not assigned to any surrogate class (samples in white), by pulling them closer to their nearest classes and pushing them further from the rest.}
\label{fig:pipeline}
\end{figure*}

To overcome the fundamental limitations of these approaches we propose to: \emph{(i)} Cast similarity learning as a surrogate classification task, using compact groups of mutually related samples as surrogates classes in a self-supervision spirit. \emph{(ii)} Combine classification with a partial ordering of samples. Even samples, which cannot be assigned to any surrogate class due to unreliable initial similarities are thus incorporated during training and in contrast to discrete classification, more fine-grained relationships are obtained due to the ordering. \emph{(iii)} Explicitly optimize similarities in a given representation space, instead of using the representation space indirectly learnt by intermediate layers of a CNN trained for classification.\emph{(iv)} Jointly optimize the surrogate classification tasks for similarity learning and the underlying grouping in a recurrent framework which is end-to-end trainable. Fig. \ref{fig:pipeline} shows a conceptual pipeline of the proposed approach.

Experimental evaluation on diverse tasks of pose estimation and object classification shows state-of-the-art performance on standard benchmarks, thus underlining the wide applicability of the proposed approach. In the pose estimation experiments we show that our method learns a general representation, which can be transferred across datasets and is even valuable for initialization of supervised methods. In addition, in the object classification experiments we successfully leverage large unlabeled datasets to learn representations in the fashion of zero-shot learning.

\section{Related Work}

Similarity learning has been a problem of major interest for the vision community from its early beginnings, due to its broad applications. With the advent of CNNs, several approaches have been proposed for supervised similarity learning using either pairs \cite{ConvNetSimPatch}, or triplets \cite{ConvNetSimTriplet} of images. Furthermore, recent works by Misra et al. \cite{shuffleandlearn}, Wang et al. \cite{ConvNetpretext2}, and Doersh et al. \cite{ConvNetpretext1} showed that temporal information in videos and spatial context information in images can be utilized as a convenient supervisory signal for learning feature representation with CNNs in an unsupervised manner. However, either supervised or unsupervised, all these formulations for learning similarities require that the supervisory information scales quadratically for pairs of images, or cubically for triplets. This results in very large training time. Furthermore, tuple and triplet formulations advocate on the CNN to indirectly learn to conceal unrelated pairs of samples (i.e. pairs that were not tied to any anchor) that are processed in different, independent batches during training. Another recent approach that has been proposed for learning similarities in an unsupervised manner is to build a surrogate (i.e. an artificial) classification task either by utilizing heavy data augmentation \cite{exemplarcnn} or by clustering based on initial weak estimates of similarities \cite{cliquecnn,pseudolabel}. The advantage of these approaches over tuple or triplet formulations is that several relationships of similarity (samples in the same class) and dissimilarity (samples in other classes) between samples are utilized during training. This results in more efficient training procedures, avoiding to sample millions of pairs or triplets of samples and encoding richer relationships between samples.

In addition, similarity learning has also been studied from the perspective of metric learning approaches \cite{wsabie,nca,magnet}. In the realm of supervised metric learning methods, Roweis et. al \cite{nca} formulated metric learning as a cross-entropy based classification problem in which all pair-wise neighbouring samples are pulled together while non-neighbouring samples are pushed away. However, provided that clusters of neighbouring points can have an arbitrary large number of samples, this strategy fails to scale to the large image collections used for unsupervised learning of similarities. Further efforts \cite{nca_ext1,nca_ext2} have tried to reduce the computational cost of performing all pairwise comparisons \cite{nmc}. \textcolor{black}{Recently, \cite{lowdensity} leveraged low-density classifiers to enable the use of large volumes of unlabelled data during training. However, \cite{lowdensity} cannot be successfully applied to the unsupervised scenario, since it requires a strongly supervised initialization , e.g. an ImageNet pre-trained model.}

\section{Approach}

In this section we show how to combine partially ordered sets (posets) of samples and surrogate classification to learn fine-grained similarities in an unsupervised manner. Key steps of the approach include: \emph{(i)} Compute compact groups of mutually related samples and use each group as a surrogate class in a classification task. \emph{(ii)}  Learn fine-grained similarities by modelling partial orderings to also leverage those samples that cannot be assigned to a surrogate class. \emph{(iii)} Due to the interdependence of grouping and similarity learning we jointly optimize them in a recurrent framework. Fig. \ref{fig:pipeline} shows a visual example of the main steps of our approach.

% Approaches for supervised similarity learning \cite{} have used all pairwise relationships of samples in a class during the optimization. Naturally, the shear number of these pairwise assignments impede \cite{} to scale to the large-scale volumes of data used for unsupervised learning. 

\subsection{Grouping}\label{sec:clustering}

To formulate unsupervised similarity learning as a classification approach we need to define surrogate classes, since labels are not available. To compute these surrogate classes we first gather compact groups of samples using standard feature distances (LDA whitened HOG \cite{hoglda,josepr,angelacvpr14}). HOG-LDA is a computationally effective foundation for estimating similarities between a large number of samples. Let our training set be defined as $\X \in \mathbb{R}^{n \times p}$, where $n$ is the total number of samples and $\x_i$ is the $i-$th sample. Then, the HOG-LDA similarity between a pair of samples $\x_i$ and $\x_j$ is defined as $s_{ij}=\exp(-\| \phi(\x_i) - \phi(\x_{j})\|_2)$. Here $\phi(\x_i) \in \mathbb{R}^{1 \times d}$ is the $d-$dimensional representation of sample $\x_i$ in the HOG-LDA feature space.

Albeit unreliable to relate all samples to another, HOG-LDA similarities can be used to find the nearest and furthest neighbors, as highly similar and dissimilar samples to a given anchor sample $\x_i$ stand out from the similarity distribution. Therefore, to build surrogate classes (i.e. compact groups of samples) we group each $\x_i$ with its immediate neighborhood (samples with similarity within the top $5\%$)
%apply a complete-linkage clustering
%starting at each $\x_i$ to merge the sample with its immediate neighborhood, 
so that all merged samples are mutually similar. These groups are compact, differ in size, and may be mutually overlapping. To reduce redundancy, highly overlapping classes are subsequently merged by agglomerative clustering, which terminates if intra-class similarity of a surrogate class is less than half of its constituents. We denote the set of samples assigned to the $c$-th surrogate class as $\mathcal{C}_c$, and the label assigned to each sample as $\y \in \{-1, 0,\dots, C-1\}^{1 \times n}$, where the label assigned to sample $\x_i$ is denoted as $y_i$. All samples that are not assigned to any surrogate class get label $-1$.

% Assigning the same label to all the nearest and another label to all the furthest neighbors of an exemplar is inappropriate. The samples in these groups may be close to $\d_i$ (or distant for the negative group) but not to another due to lacking transitivity.

\subsection{Partially Ordered Sets}

\begin{figure}[!t]
\includegraphics[width=0.46\textwidth, height=3 cm]{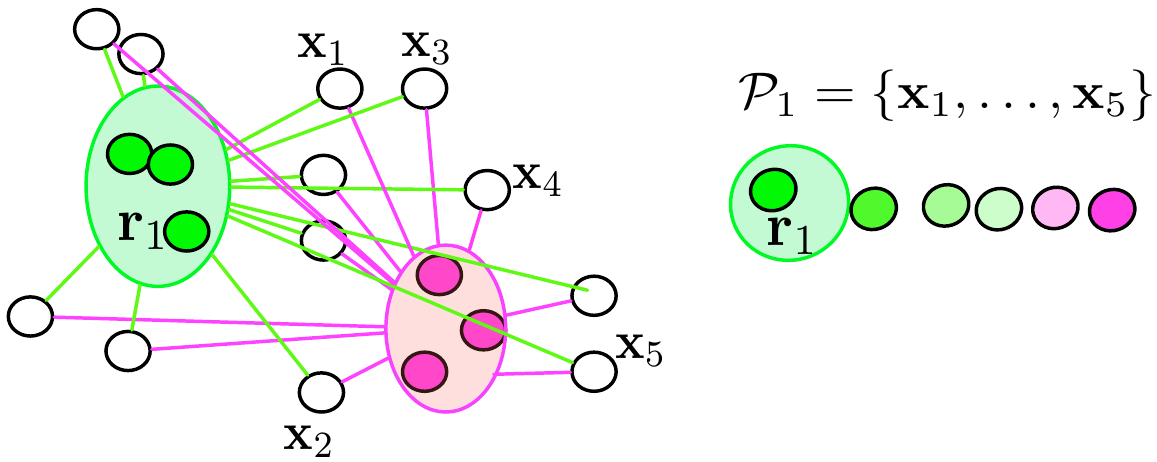}
\caption{Visual interpretation of a poset. Samples assigned to a surrogate class are shaded in a particular color, while samples not assigned to surrogate classes are represented in white.}
\label{fig:poset}
\end{figure}

Provided the unreliability of similarity estimates used for building surrogate classes, a large number of samples cannot be assigned to any class, because they are neither similar nor dissimilar to any sample. This deprives the optimization of using all available data during training. As a result, fine-grained similarities are poorly represented, since learning to classify surrogate classes does not model relative similarities of samples that are not assigned to any class. To overcome this limitation we leverage the information encoded in posets of samples relative to a surrogate class. That is, for each sample not ~assigned to any surrogate class (i.e. $\x_i~:~y_i~=~-1$) we compute a soft assignment (i.e. a similarity score) to the $Z$ nearest surrogate classes $\mathcal{C}_z : z \in \{1, \dots, Z\}$. Once all unlabeled points are softly assigned to their $Z$ nearest classes, we obtain as a result, a poset $\mathcal{P}_c$ for each class. Thus, a poset $\mathcal{P}_c$ is a set of samples which are softly assigned to class $\mathcal{C}_c$. Posets can be of variable size and partially overlapping. We show a visual example of a poset in Fig. \ref{fig:poset}.

%ToDo: the last two sentences are unclear

Formally, given a deep feature representation $\phi^\theta$ (e.g an arbitrary layer in a CNN with parameters $\theta$), and a surrogate class $\mathcal{C}_c$, a poset of unlabeled samples $\mathcal{P}_c~=~\{\x_j,\dots, \x_k\}~:~y_j~=~y_k~=~-1~\forall~j,k$  with respect to $\mathcal{C}_c$ is defined as:
\begin{alignat}{2}\label{eq:partial_order}
&\forall_{\x_i \in \mathcal{C}_c} \{ \exp(-\| \phi^\theta(\x_i) - \phi^\theta(\x_j) \|_2) > \nonumber \\ 
&\exp (-\| \phi^\theta(\x_i) - \phi^\theta(\x_k) \|_2) \} \iff j < k \forall j,k.
\end{alignat}

In Eq. \eqref{eq:partial_order} a poset is defined by computing the similarity of unlabeled sample $\x_j$ to all the samples in class $\mathcal{C}_c$, which during training is costly to optimize. However, due the compactness of our grouping approach, which only gathers very similar samples into surrogate $\mathcal{C}_c$, we can effectively replace the similarities to all points in $\mathcal{C}_c$ by the similarity to a representative sample $\bar{\x}_c$ in $\mathcal{C}_c$, which is the class medioid, $ \bar{\x}_c = \argmin\limits_{\x_i \in \mathcal{C}_c} \sum\limits_{\x_j \in \mathcal{C}_c} \|\phi^\theta(\x_i) - \phi^\theta(\x_j) \|_2$. 

Following the definition of a poset in Eq. \ref{eq:partial_order}, the widely adopted tuple and triplet formulations \cite{ConvNetpretext1, ConvNetpretext2,shuffleandlearn,Yang_2016_CVPR} are a specific case of a poset in which $\mathcal{P}$ contains at most $2$ samples, and $\mathcal{C}_c$ contains just one. In this sense, deep feature representations $\phi$ (i.e. CNNs) trained using triplet losses seek to sort two pairs of samples (i.e. anchor-positive and anchor-negative) according to their similarity. As a result, triplet formulations rely on the CNN to \textit{indirectly} learn to compare and reconcile the vast number of \textit{unrelated} sampled pairs that were processed on different, independent mini-batches during training. In contrast, posets, explicitly encode an ordering between a large number of sample pairs (i.e pairs consisting of an unlabeled sample and its nearest class representative). Therefore, using posets during training enforces the CNN to order all unlabeled samples $\x_i~:~y_i~=~-1$ according to their similarity to the $Z$ nearest class representatives $\br^z_i~:z~\in~\{1,~\dots~,~Z\}$, where $\br^z_i$ is the $z-$th nearest $\bar{\x}_c$ to sample $\x_i$, learning fine-grained interactions between samples.  Posets generalize tuple and triplet formulations by encoding similarity relationships between unlabeled samples to make a decision whether to move closer to a surrogate class. This effectively increases our training set when compared to just using the samples assigned to surrogate classes, and allows us to model finer relationships.

% I think its clear now that what we achieved cannot be obtained by sampling just more triplets.

%ToDo: This paragraph seems somewhat critical. Aren't these just the normal triplet constraints and we also use a lot of indepently obtained constraints separetely and hope the cnn balances them? In other words, doesn't the critique before also apply here?

\subsection{Objective function}

In our formulation, we strive for a trade-off model in which we jointly optimize a surrogate classification task and a metric loss to capture the fine-grained similarities encoded in posets. Therefore, we seek an objective function $\mathcal{L}$ which penalizes: \emph{(i)} misclassifications of samples $\x_i$ with respect to their surrogate label $y_i$, and \emph{(ii)} similarities of samples $\x_i~:~y_i~=-1$. with respect to their $Z$ nearest class representatives. The objective function should inherit the reliability of framing similarity learning as surrogate classification tasks, while using posets to incorporate those training samples that were previously ignored because they could not be assigned to any surrogate class. In particular, we require the CNN to pull samples from posets $\x_i~\in~\mathcal{P}_c$ closer to their $Z$ nearest class representatives, while pushing them further from all other class representatives in a training mini-batch.  Furthermore, we require that unreliable similarities (i.e. samples that are far from all surrogate classes), vanish from the loss, rendering the learning process robust to outliers. In addition, in order to capture fine-grained similarity relationships, we want to directly optimize the feature space $\phi$ in which similarities are computed.

Therefore, let $\R^z \in \mathbb{R}^{n \times d}$ denote the $z$-th nearest class representatives of each unlabeled sample $\x_i~:~y_i~=~-1$, where $\br^z_i$ is the $z$-th nearest class representative of sample $\x_i$, and $\theta$ be the parameters of the CNN. Then, our objective function combines the surrogate classification loss $\mathcal{L}_1$ with our poset loss $\mathcal{L}_2$:
\begin{alignat}{2}
&\mathcal{L}(\x_i, y_i, \R; \theta) = \frac{1}{N} \sum_{i=1}^{N} \mathcal{L}_1(\x_i, y_i) + \lambda\mathcal{L}_2(\x_i, \R, \phi) \label{eq:loss},
\end{alignat}
where $\lambda$ is a scalar and,
\begin{alignat}{2}
&\mathcal{L}_1(\x_i,y_i; \theta)=-\log\frac{\exp(t^{\theta}_{i,y_i})}{\sum_{j=0}^{C-1}\exp(t^{\theta}_{i,j})} \mathbbm{1}_{y_i\neq-1}, \label{eq:loss1}\\
\begin{split}
&\mathcal{L}_2(\x_i,\R; \theta)=\\
&=-\log\frac{\sum\limits_{z=1}^{Z}\exp({\frac{-1}{2\sigma^2} ( \|\ \phi^\theta(\x_i) - \phi^\theta(\br^z_i) \|_2^2-\gamma}))} {\sum_{j=1}^{C'}\exp(\frac{-1}{2\sigma^2}\|\ \phi^\theta(\x_i) - \phi^\theta(\br_j) \|_2^2)}. \label{eq:loss2}
\end{split}
\end{alignat}

In Eq. \eqref{eq:loss1}, $\t^\theta_i = \t^\theta(\x_i)$ are the logits of sample $\x_i$ for a CNN with parameters $\theta$. In Eq. \eqref{eq:loss2} $C'$ is the number of surrogate classes in the batch, $\sigma$ is the standard deviation of the current assignment of samples to surrogate classes, and $\gamma$ is the margin between surrogate classes. It is note-worthy that Eq. \eqref{eq:loss2} can scale to an arbitrary number of classes, since it does not depend on a fixed-sized output target layer, avoiding the shortcomings of large output spaces in CNN learning \cite{outputspace} \footnote{In our experiments we successfully scaled the output space to $20$K surrogate classes.}.

\begin{figure}
\includegraphics[width=0.47\textwidth]{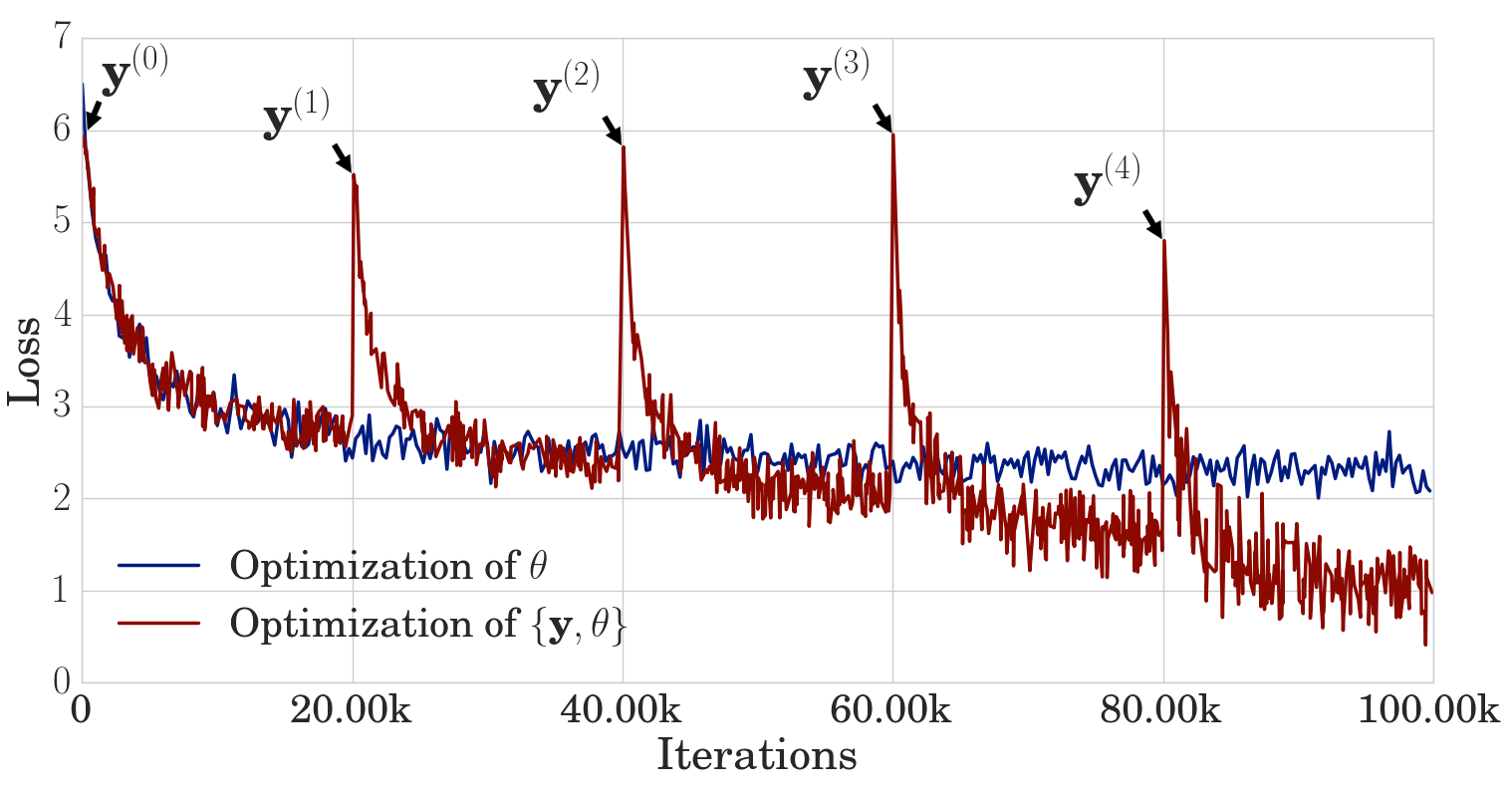}
\caption{Loss value $\mathcal{L}$ for long jump category over each unrolling step. Evidently the model benefits from jointly optimizing $\{\y, \theta\}$.}
\label{fig:loss_reclustering}
\end{figure}

Finally, note that if $Z=1$ the problem reduces to a cross-entropy based classification, where the standard logits (i.e. outputs of the last layer) are replaced by the similarity to the surrogate class representative in feature space $\phi$. However, for $Z>1$ relative similarities between surrogate classes enter into play and posets encoding fine-grained interactions naturally arise (cf. Fig. \ref{fig:interp}). In all our experiments we set $Z>=2$. During training, CNN parameters $\theta$ are updated by error-backpropagation with stochastic mini-batch gradient descent. In typical classification scenarios the training set is randomly shuffled to avoid biased gradient computations that hamper the learning process. Therefore, at training time we build our mini-batches of samples by selecting a random set of samples not assigned to a surrogate class $\x_i~:~y_i=-1$, and retrieving all the surrogate classes $\mathcal{C}_c$ which contain $\x_i$ in their poset $\x_i \in \mathcal{P}_c$. In Fig. \ref{fig:loss_reclustering} we take as a study case the \textit{long jump} category of the Olympic Sports dataset (cf. Sec. \ref{sec:experiments}) and show the $\mathcal{L}$ decreases along iterations. In particular, we show that if $\y$ and $\theta$ are optimized jointly we attain better performance.
 
\subsection{Joint Optimization}\label{sec:recurrent}

In our setup, the grouping and similarity learning tasks are mutually dependent on each other. Therefore, we strive to jointly learn a representation $\phi^\theta$, which captures similarity relationships, and an assignment of samples to surrogate classes $\y$.
A natural way to model such dependence in variables is to use a Recurrent Neural Network (RNN) \cite{rnn}. In particular, RNNs have shown a great potential to model relationships on sequential problems, where each prediction depends on previous observations. Inspired by this insight, we employ a recurrent optimization technique. Following the standard process for learning RNNs we jointly learn $\{\y, \theta\}$ by unrolling the optimization into steps. At time step $m$ we update $\y$ and $\theta$ as follows: 
\begin{alignat}{2}
\begin{split}
&\y^{(m)}=\argmax_{\y}\mathcal{G}(\X; \phi^{\theta^{(m-1)}},\y^{(m-1)})\\
&\text{s.t.}  \sum\limits_{i:y_i=c}^{n} 1 > t, \forall_{c \in \{0, \dots, C-1\} },
\label{eq:g}
\end{split}
\\
&\theta^{(m)} =\argmin\limits_{\theta} \mathcal{L}(\X, \y^{(m)}, \R^{(m)}; \theta^{(m-1)}). \label{eq:theta}
\end{alignat}
Where $\mathcal{G}$ is a cost function of pairwise clustering that favors compactness based on sample similarities, which are entailed by the representation $\phi^{\theta^{(m-1)}}$\textcolor{black}{, and $t$ is a lower bound on the number of samples of each cluster}.
%It uses representation $\phi^{\theta^{m-1}}$ and assignment $\y^{m-1}$ to compute the new assignment $\y^m$. It is defined as
%
\begin{alignat}{2}
\begin{split}
&\mathcal{G}(\X;\phi^{\theta},\y) = \\
&=\sum\limits_{c=0}^{C-1}\sum\limits_{i:y_i=c}^{n} \frac{\sum\limits_{j: y_j=c}^{n} \exp(-\| \phi^\theta(\x_i) - \phi^\theta(\x_j) \|_2)}{\left(\sum\limits_{j: y_j=c}^{n} 1\right)^2}.
\end{split}\label{eq:clustering}
\end{alignat}
\textcolor{black}{In order to avoid the trivial solution of assigning a single sample to each cluster we initialize $\y^{(0)}$ with the grouping introduced in Sec. \ref{sec:clustering} using HOG-LDA as our initial $\phi$. In our implementation, $\y$ follows a relaxed one-hot encoding, which can be interpreted as an affinity of samples to clusters. Then, Eq. \eqref{eq:g} becomes differentiable and is optimized using SGD.} Subsequently, $\mathcal{L}$ learns a deep similarity encoding representation $\phi^{\theta{(m)}}$ on samples $\X$ using assignments $\y^{(m)}$ and partial orders of $\X$ with respect to representatives $\R^{(m)}$. In a typical RNN scenario, for each training iteration the RNN is unrolled $m$ steps. However, this would be inefficient in our setup, as the CNN representation $\phi^\theta$ is learnt using SGD, and thus, requires to be optimized for a large number of iterations to be reliable, especially at the first unrolled steps. Therefore, at each step $m$, we find $\theta^{(m)}$ by optimizing Eq. \eqref{eq:theta} for a number of iterations, fixing $\y^{(m)}$ and $\R^{(m)}$. Then, we use $\theta^{(m)}$ to find the optimal $\y^{(m+1)}$ by optimizing $\mathcal{G}$ using SGD. \textcolor{black}{The presented RNN can also be interpreted as block-coordinate descent \cite{blockcd}, where the grouping $\y$ is fixed while updating the representation parameters $\theta$ and vice versa. The convergence of block coordinate-descent methods has been largely discussed obtaining guarantees of convergence to a stationary point \cite{convergence,convergence2}.}

\section{Experiments}\label{sec:experiments}

In this section we present a quantitative and qualitative analysis of our poset based approach on the challenging and diverse scenarios of human pose estimation and object classification. In all our experiments we adopt the AlexNet architecture \cite{alexnet}.

\subsection{Human Pose Estimation}

To evaluate the proposed approach in the context of pose estimation we consider $3$ different datasets, Olympic Sports (OS), Leeds Sports Pose (LSP), and MPII-Pose (MPI). We show that our unsupervised method is valuable for a range of retrieval problems: For OS we evaluate zero-shot retrieval of detailed postures. On LSP, we perform zero-shot and semi-supervised estimation of pose. Finally, on MPII we evaluate our approach as an initialization for a supervised learning approach for pose estimation. In contrast to other methods that fine-tune supervised initializations of a CNN, we train our AlexNet \cite{alexnet} architecture from scratch.

\subsubsection{Olympic Sports}\label{sec:olympic_metrics}

The Olympic Sports dataset \cite{olympic_sports} is a compilation of video sequences of different $16$ sports competitions, containing more than $110000$ frames overall. We use the approach of \cite{dpm} to compute person bounding boxes and utilize this large dataset to learn a general representation that encodes fine-grained posture similarities. In order to do so, we initially compute $20000$ surrogate classes consisting of $8$ samples in average. Then, we utilize partially ordered sets of samples not assigned to any surrogate classes. To train our RNN we use the optimization approach described in Sec. \ref{sec:recurrent}, where the RNN is unrolled on $m=10$ steps. At each unrolled step, $\theta$ is updated during $20000$ iterations of error-backpropagation. To evaluate our representation on fine-grained posture retrieval we utilize the annotations provided by \cite{cliquecnn} 
on their project webpage\footnote{\label{fn:cliquecnn} \url{https://asanakoy.github.io/cliquecnn/}} and follow their evaluation protocol, using their annotations only for testing. \textcolor{black}{We compare our method with CliqueCNN \cite{cliquecnn} by directly evaluating their models provided at \textsuperscript{\normalfont{\ref{fn:cliquecnn}}}}, the triplet formulation of Shuffle\&Learn \cite{shuffleandlearn}, the tuple approach of Doersch et. al \cite{ConvNetpretext1}, Exemplar-CNN \cite{exemplarcnn}, Alexnet \cite{alexnet}, Exemplar-SVMs \cite{exemplarsvm}, and HOG-LDA \cite{hoglda}. For completeness we also include a version of our model that was initialized with Imagenet model \cite{alexnet}. 
During training we use as $\phi$ the \emph{fc7} output representation of Alexnet and compute similarities using cosine distance. We use \emph{Tensorflow} \cite{tensorflow} for our implementation. (\emph{i}) For CliqueCNN, Shuffle\& Learn, and Doersh et. al methods we use the models downloaded from their respective project websites. \emph{(ii)} Exemplar-CNN is trained using the best performing parameters reported in \cite{exemplarcnn} and the 64c5-128c5-256c5-512f architecture. Then we use the output of fc4 and compute 4-quadrant max pooling. \emph{(iii)} Exemplar-SVM was trained on the exemplar frames using the HOG descriptor. The samples for hard negative mining come from all categories except the one that an exemplar is from. We performed cross-validation to find an optimal number of negative mining rounds (less than three). The class weights of the linear SVM were set as $C_1=0.5$ and $C_2=0.01$. During training of our approach, each image in the training set is augmented by performing random translation, scaling and rotation to improve invariance with respect to these.

In Tab. \ref{tab:avg_auc} we show the average AuC over all categories for the different methods. When compared with the best runner up \cite{cliquecnn}, the proposed approach improves the performance $2\%$ (the method in \cite{cliquecnn} was pre-trained on Imagenet). This improvement is due to the additional relationships established by posets on samples not assigned to any surrogate class, which \cite{cliquecnn} ignored during training. In addition, when compared to the state-of-the-art methods that leverage tuples \cite{ConvNetpretext1} or triplets \cite{shuffleandlearn} for training a CNN from scratch, our approach shows $16\%$ higher performance. This is explained by the more detailed similarity relationships encoded in each poset, which in tuple methods the CNN has to learn implicitly.

% \begin{table}
%     \centering
%     \begin{tabular}{|c|c|c|c|}
%     \hline
%     HOG-LDA \cite{hoglda} & Ex-SVM \cite{exemplarsvm} & Ex-CNN \cite{exemplarcnn} & Alexnet \cite{alexnet} \\ \hline
%     0.62 &  0.72  &   0.64   & 0.65 \\ \hline
%     \hline
%     Doersch et. al \cite{ConvNetpretext1} & Suffle\&Learn \cite{shuffleandlearn}  & CliqueCNN \cite{cliquecnn} & Ours scratch/Imgnet \\ \hline
%      0.62 & 0.63 & 0.83 & \textbf{0.78/0.85} \\ \hline
%     \end{tabular}
%     \caption{Avg. AUC for each method on Olympic Sports dataset.}
%     \label{tab:avg_auc}
% \end{table}

\begin{table}
    \scriptsize
    \centering
    \begin{tabular}{|c|c|c|}
    \hline
    HOG-LDA \cite{hoglda} & Ex-SVM \cite{exemplarsvm} & Ex-CNN \cite{exemplarcnn}\\ \hline 
    0.62 &  0.72  &   0.64   \\ \hline \hline
    Alexnet \cite{alexnet} & Doersch et. al \cite{ConvNetpretext1} & Suffle\&Learn \cite{shuffleandlearn}   \\ \hline
    0.65 & 0.62 & 0.63 \\ \hline \hline
    CliqueCNN \cite{cliquecnn} & Ours scratch &  Ours Imagenet\\ \hline
      0.83 & 0.78 & \textbf{0.85} \\ \hline
    \end{tabular}
    \caption{Avg. AUC for each method on Olympic Sports dataset.}
    \label{tab:avg_auc}
\end{table}

\begin{figure*}[!t]
\includegraphics[width=0.99\textwidth, height= 4 cm]{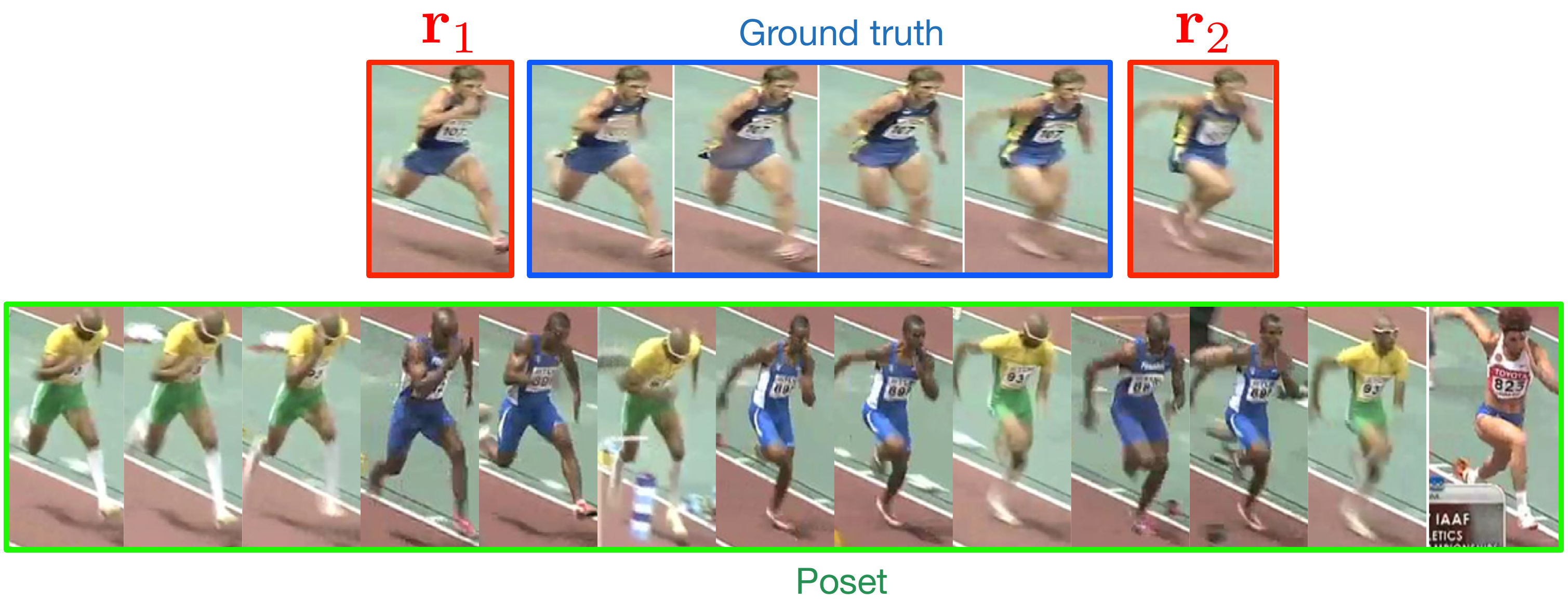}
\caption{Partially ordered set learnt by the proposed approach. The top row shows two surrogate class representatives (highlighted in red) of the same sequence and the ground truth sub-sequence between them highlighted in blue. The bottom row shows the predicted poset highlighted in green, successfully capturing fine-grained similarities.}
\label{fig:interp}
\end{figure*}

In addition to the quantitative analysis we also perform a qualitative evaluation of the similarities learnt by the proposed method. In order to do so, we take a sequence from the \textit{long jump} category of Olympic Sports and select two representatives $\{\br_1, \br_r\}$ with a gap of $8$ frames between them and show in Fig. \ref{fig:interp} the poset learnt by our approach. The top row shows two representatives of the same sequence highlighted in red and the remaining sub-sequence between them in blue. In the bottom row, we present the poset learnt by our approach.
Since $\br_1$ and $\br_2$ show different parts of a short gait cycle, the similarity relations in the poset should set other frames into perspective and order them.
%close in time (0.3s) the natural course of similarities between $\br_1$ and $\br_2$ has to encode temporal coherence.
And indeed, we observe that the poset successfully encodes this temporal coherence by ordering frames from other sequences that fit in this gap. This is even more interesting, since during training absolutely no temporal structure was introduced in the model, as we were training on only individual frames. These results spurred our interest to also apply the learnt posets for video reconstruction using only few sparse representatives per sequence, additional results can be found in the supplementary material.

\subsubsection{Leeds Sports Pose}

\begin{figure*}
\includegraphics[width=\textwidth, height=4 cm]{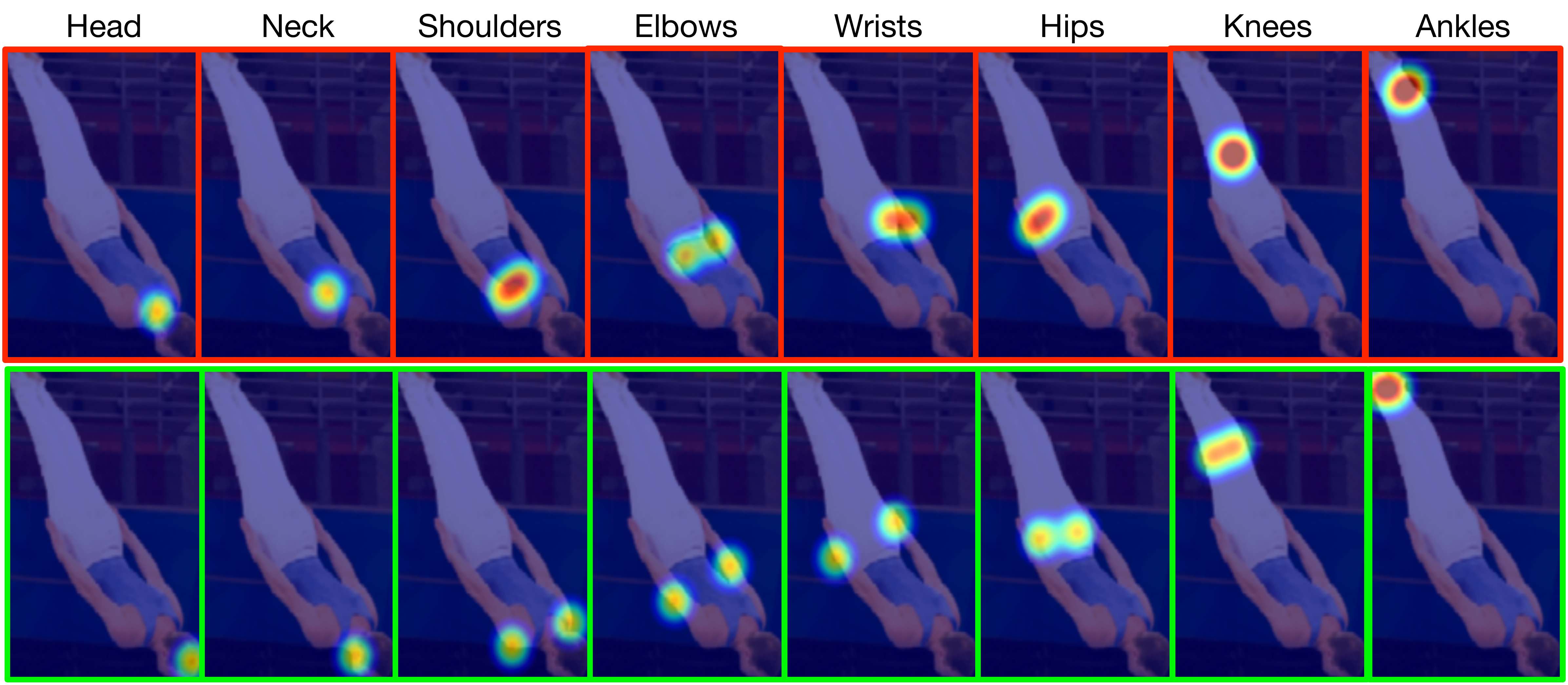}
\caption{Top row: Heatmaps obtained by DeepPose (stg-1) \cite{deeppose} trained on LSP, highlighted in red. Bottom row: Heatmaps obtained by our zero-shot unsupervised approach, highlighted in green.}
\label{fig:heatmap_pred}
\end{figure*}

After evaluating the proposed method for fine-grained posture retrieval, we tackle the problem of zero-shot pose estimation on the LSP dataset. That is, we transfer the pose representation learnt on Olympic Sports to the LSP dataset and retrieve similar poses based on their similarity. The LSP \cite{lsp} dataset is one of the most widely used  benchmarks for pose estimation. In order to evaluate our model we then employ the fine-grained pose representation learnt by our approach on OS, and transfer it to LSP, without doing any further training. For evaluation we use the representation to compute visual similarities and find nearest neighbours to a query frame. Since the evaluation is zero-shot, joint labels are not available. At test time we therefore estimate the joint coordinates of a query person by finding the most similar frame from the training set and taking its joint coordinates. We then compare our method with Alexnet \cite{alexnet} pre-trained on Imagenet, the triplet approach of Misra et. al (Shuffle\&Learn) \cite{shuffleandlearn} and CliqueCNN \cite{cliquecnn}. In addition, we also report an upper bound on the performance that can be achieved by zero-shot evaluation using ground-truth similarities. Here the most similar pose for a query is given by the frame, which is closest in average distance of ground-truth pose annotations. This is the best one can achieve without a parametric model for pose (the performance gap to $100\%$ shows the discrepancy between poses in test and train set).
%that there are test poses for which there is no nearest neighbor in train set with identical pose).
%do by finding the most similar frame, when not provided with a supervised parametric model (the performance gap to $100\%$ shows the difference between training and test poses).
For completeness, we compare with a fully supervised state-of-the-art approach for pose estimation \cite{posemachines}. For computing simialarities we use the same experimental settings described in Sect. \ref{sec:olympic_metrics}, where $\phi$ is the representation extracted from \emph{pool5} layer of Alexnet. In Tab. \ref{tab:results_lsp} we show the PCP@$0.5$ obtained by the different methods. For a fair comparison with CliqueCNN \cite{cliquecnn} (which was pre-trained on Imagenet), we include a version of our method trained using Imagenet initialization. Our approach significantly improves the visual similarities learned using both Imagenet pre-trained AlexNet and CliqueCNN \cite{cliquecnn}, obtaining a performance boost of at least $4\%$ in PCP score. In addition, when trained from scratch without any pre-training on Imagenet our model outperforms the recent triplet model of \cite{shuffleandlearn} by $4\%$, due to the fact that posets are a natural generalization of triplet models, which encode finer relationships between samples. Finally, it is notable that even though our pose representation is \emph{transferred from a different dataset} without fine-tuning on LSP, it obtains state-of-the-art performance. In Fig. \ref{fig:heatmap_pred} we show a qualitative comparison of the part predictions of the supervised approach in \cite{deeppose} trained on LSP, with the heatmaps yielded by our zero-shot approach. 

\begin{table}[!t]
    \scriptsize
    \centering
    \begin{tabular}{|c|c|c|c|c|c|c|c|}
    \hline
    Method  &T &UL &LL &UA &LA &H  &Total \\
    \hline
    Ours - Imagenet & 83.5 & 54.0 & 46.8 & 34.1 & 16.8 & 54.3 &  \textbf{48.3}\\
    \hline
    CliqueCNN \cite{cliquecnn}  & 80.1 & 50.1 & 45.7 & 27.2 & 12.6 &  45.5 &  43.5 \\
    \hline
    Alexnet\cite{alexnet}& 76.9 & 47.8 & 41.8 & 26.7 & 11.2 & 42.4 & 41.1 \\
    \hline
    \hline
    Ours - Scratch & 67.0 & 38.6 & 34.9 & 20.5 & 9.8 & 35.1 &  \textbf{34.3}\\
    \hline
    Shuffle\&Learn \cite{shuffleandlearn}  & 60.4  & 33.2 & 28.9 & 16.8 & 7.1 &  33.8 &  30.0\\
    \hline
    \hline
    Ground Truth & 93.7  & 78.8 & 74.9 & 58.7 & 36.4 & 72.4 & 69.2\\
    \hline
    \hline
    P. Machines \cite{posemachines} & 93.1 & 83.6 & 76.8 & 68.1 & 42.2 & 85.4 & 72.0 \\
    \hline
    \end{tabular}
    \caption{PCP measure for each method on Leeds Sports dataset for zero-shot pose estimation.}
    \label{tab:results_lsp}
\end{table}

In addition to the zero-shot learning experiments we also used our pose representation learnt on Olympic Sports as an initialization for learning the DeepPose method \cite{deeppose} on LSP in a semi-supervised fashion. Our implementation of this method is available at github\footnote{\url{https://github.com/asanakoy/deeppose_tf}}. To evaluate the validity of our representation we compare the performance obtained by DeepPose \cite{deeppose}, when trained with one of the following models as initialization: random initialization, Shuffle\&Learn \cite{shuffleandlearn} (triplet model), and our approach trained on OS. For completeness, we also compared with Imagenet pre-trained AlexNet \cite{alexnet}. Tab. \ref{tab:results_lsp_deeppose} shows the PCP@$0.5$ obtained by training DeepPose (stg-1) using their best reported parameters. The obtained results show that our representation successfully encodes pose information, obtaining a performance boost of $9\%$ when compared with a random initialization (that our model starts from), since we learn general pose features that act as a regularizer during training. A note-worthy comparison is that the difference between utilizing Imagenet pre-training, which uses $1.2$ million labeled images, and our unsupervised learning approach is just $5\%$.

\begin{table}[!t]
    \scriptsize
    \centering
    \begin{tabular}{|c|c|c|c|c|c|c|c|}
    \hline
    Initialization  &T &UL &LL &UA &LA &H  &Total \\
    \hline
    Ours & 89.7 & 62.1 & 48.2 & 36.0 & 16.0 & 54.2 &  51.0\\
    \hline
    Shuffle\&Learn \cite{shuffleandlearn} & 90.4 & 62.7 & 45.7 & 33.3 & 11.8 & 52.0 &  49.3\\
    \hline
    Random init. & 87.3 & 52.3 & 35.4 & 25.4 & 7.6 & 44.0 &  42.0\\
    \hline
    \hline
    Alexnet \cite{alexnet} & 92.8 & 68.1 & 53.0 & 39.8 & 17.5 & 62.8 &  55.7\\
    \hline
    \end{tabular}
    \caption{PCP measure for each method on Leeds Sports dataset using different methods as initialization for the DeepPose method \cite{deeppose}.}
    \label{tab:results_lsp_deeppose}
\end{table}

\subsubsection{MPII Pose}

We now evaluate our approach in the challenging MPII Pose dataset \cite{mpii} which is a state of the art benchmark for evaluation of articulated human pose estimation. The dataset includes around $25$K images containing over $40$K people with annotated body joints. MPII Pose is a particularly challenging dataset because of the clutter, occlusion and number of persons appearing in images. To evaluate our approach in MPII Pose we follow the semi-supervised training protocol used for LSP and compare the performance obtained by DeepPose \cite{deeppose}, when trained using as initialization each of the following models: Random initialization, Shuffle\&Learn \cite{shuffleandlearn} (triplet model) and our approach trained on OS. For completion, we also evaluate Imagenet pre-trained AlexNet \cite{alexnet} as initialization. Following the standard evaluation metric on MPII dataset, Tab. \ref{tab:results_mpii_deeppose} shows the PCKh@$0.5$ obtained by training DeepPose (stg-1) using their best reported parameters with the different initializations.

The performance obtained on MPII Pose benchmark shows that our unsupervised representation successfully scales to challenging datasets, successfully dealing with clutter, occlusions and multiple persons. In particular, when comparing our unsupervised initialization with a random initialization we obtain a $7\%$ performance boost, which indicates that our features encode a robust notion of pose that is robust to the clutter present in MPII dataset. Furthermore, we obtain a $3\% $ improvement over the  Shuffle\&Learn \cite{shuffleandlearn} approach, due to the finer-grained relationships encoded by posets. Finally, it is important to note that the difference between utilizing Imagenet pre-trained AlexNet\cite{alexnet}, and our unsupervised learning approach is just $5\%$.

\begin{table}[!t]
    \scriptsize
    \centering
    \begin{tabular}{|c|c|c|c||c|}
    \hline
    & Ours & Shuffle\&Learn \cite{shuffleandlearn} &  Random Init.  & AlexNet\cite{alexnet} \\
    \hline
    Head & 83.8 & 75.8 & 79.5 & 87.2 \\
    \hline
    Neck  & 90.9 & 86.3 & 87.1 & 93.2 \\
    \hline
    LR Shoulder & 77.5 & 75.0 & 71.6 & 85.2\\
    \hline
    LR Elbow. & 60.8 & 59.2 & 52.1 & 69.6\\
    \hline
    LR Wrist  & 44.4 & 42.2 & 34.6 & 52.0\\
    \hline
    LR Hip & 74.6 & 73.3 & 64.1 & 81.3\\
    \hline
    LR Knee & 65.4 & 63.1 & 58.3 & 69.7\\
    \hline
    LR Ankle & 57.4 & 51.7 & 51.2 & 62.0\\
    \hline
    Thorax & 90.5 & 87.1 & 85.5 & 93.4\\
    \hline
    Pelvis & 81.3 & 79.5 & 70.1 & 86.6\\
    \hline
    \hline
    Total & 72.7 & 69.3 & 65.4 & 78.0 \\
    \hline
    \end{tabular}
    \caption{PCKh@$0.5$ measure for each initialization method on MPII Pose benchmark dataset using different initializations for the DeepPose approach \cite{deeppose}.}
    \label{tab:results_mpii_deeppose}
\end{table}

\subsection{Object Classification on PASCAL VOC}
To evaluate the general applicability of our approach, let us now switch from human pose estimation to the challenging diverse problem of object classification.
%to be applied in different scenarios, we evaluate it in PASCAL VOC object classification \cite{voc}.
%\subsubsection{PASCAL VOC}
%
We classify object bounding boxes of the PASCAL VOC 2007 \cite{voc} dataset in zero-shot fashion by predicting the most similar images to a query. The object representation needed for computing similarities, we obtain without supervision information, using visual similarities of the triplet model of Wang et al. \cite{ConvNetpretext2} as initializiation. Neither this initialization nor our method apply pre-training or fine tuning on ImageNet or Pascal VOC. Using this initialization we then compute an initial clustering on  $1000$ surrogate classes with $8$ samples in average, on the training set images. We then utilize partially ordered sets of samples not assigned to any class, and jointly optimize assignments and representation using the recurrent optimization approach describe in Sec. \ref{sec:recurrent}. The representation $\phi$ used to compute similarities on the PASCAL datasets is for each CNN method that we now compare the \emph{fc6} layer.
We compare our approach with HOG-LDA \cite{hoglda}, the triplet approach of \cite{ConvNetpretext2}, CliqueCNN \cite{cliquecnn}, Imagenet pre-trained AlexNet \cite{alexnet}, and RCNN \cite{rcnn}. In Tab. \ref{tab:results_voc} we show the classification performance for all methods for $k=5$ (for $k>5$ there was only insignificant performance improvement). Our approach improves upon the initial similarities of the unsupervised triplet approach of \cite{ConvNetpretext2} to yield a performance gain of $6\%$ without requiring any supervision information or fine-tuning on PASCAL.

\begin{table}[!h]
    \scriptsize
    \centering
    \begin{tabular}{|c |c| c|}
    \hline
    HOG-LDA & Wang et. al \cite{ConvNetpretext2} & CliqueCNN\cite{cliquecnn}\\  
    \hline
    0.1180 &  0.4501 & 0.4812 \\
    \hline\hline
    Wang et.al \cite{ConvNetpretext2} + Ours & Alexnet \cite{alexnet} & RCNN \cite{rcnn}\\
    \hline
    0.5101 &0.6160 & 0.6825 \\
    \hline
    \end{tabular}
    \caption{Classification results for PASCAL VOC 2007}
    \label{tab:results_voc}
\end{table}

\section{Conclusions}
We have presented an unsupervised approach to similarity learning based on CNNs by framing it as a combination of surrogate classification tasks and poset ordering. This generalizes the widely used tuple and triplet losses to establish relations between large numbers of samples. Similarity learning then becomes a joint optimization problem of grouping samples into surrogate classes while learning the deep similarity encoding representation. In the experimental evaluation the proposed approach has shown competitive performance when compared to state-of-the-art results, learning fine-grained similarity relationships in the context of human pose estimation and object classification.\\

\noindent\textbf{Aknowledgments:} This research has been funded in part by the Heidelberg Academy of Sciences. We are grateful to the NVIDIA corporation for donating a Titan X GPU.

\bibliographystyle{plain}
\bibliography{egbib}

\end{document}